\documentclass[11pt]{article}
\usepackage{acl2015}
\usepackage{times}
\usepackage{url}
\usepackage{latexsym}
\usepackage{tikz-dependency}
\usepackage{fancyvrb}
\usepackage{caption}
\usepackage{subcaption}

\title{Dependency Recurrent Neural Language Models for Sentence Completion}

\author{Piotr Mirowski \\
  Google DeepMind \\
  {\tt piotr.mirowski@computer.org} \\\And
  Andreas Vlachos \\
  University College London \\
  {\tt a.vlachos@cs.ucl.ac.uk} \\}

\date{}

\begin{document}
\maketitle
\begin{abstract}
 Recent work on language modelling has shifted focus from count-based models to neural models. In these works, the words in each sentence are always considered in a left-to-right order. In this paper we show how we can improve the performance of the recurrent neural network (RNN) language model by incorporating the syntactic dependencies of a sentence, which have the effect of bringing relevant contexts closer to the word being predicted. We evaluate our approach on the Microsoft Research Sentence Completion Challenge and show that the dependency RNN proposed improves over the RNN by about 10 points in accuracy. Furthermore, we achieve results comparable with the state-of-the-art models on this task.
\end{abstract}

\section{Introduction}

Language Models (LM) are commonly used to score a sequence of tokens according to its probability of occurring in natural language.
 They are an essential building block in a variety of applications such as machine translation, speech recognition and grammatical error correction.
 The standard way of evaluating a language model has been to calculate its perplexity on a large corpus. However, this evaluation assumes the output of the language model to be probabilistic and it has been observed that perplexity does not always correlate with the downstream task performance.
 
For these reasons, \newcite{zweig2012challenge} proposed the Sentence Completion Challenge, in which the task is to pick the correct word to complete a sentence out of five candidates. Performance is evaluated by accuracy (how many sentences were completed correctly), thus both probabilistic and non-probabilistic models (e.g. \newcite{roark2007lm}) can be compared. Recent approaches for this task include both neural and count-based language models \cite{zweig2012computational,gubbins2013lm,mnih2013lblnce,mikolov2013efficient}.

Most neural language models consider the tokens in a sentence in the order they appear, and the hidden state representation of the network is typically reset at the beginning of each sentence.
In this work we propose a novel neural language model that learns a recurrent neural network (RNN) \cite{Mikolov:10} on top of the syntactic dependency parse of a sentence. Syntactic dependencies bring relevant contexts closer to the word being predicted, thus enhancing performance as shown by \newcite{gubbins2013lm} for count-based language models.
Our Dependency RNN model is published simultaneously with another model, introduced in \newcite{tai2015}, who extend the Long-Short Term Memory (LSTM) architecture to tree-structured network topologies and evaluate it at sentence-level sentiment classification and semantic relatedness tasks, but not as a language model.

  Adapting the RNN to use the syntactic dependency structure required to reset and run the network on all the paths in the dependency parse tree of a given sentence, while maintaining a count of how often each token appears in those paths.
  Furthermore, we explain how we can incorporate the dependency labels as features.

Our results show that the dependency RNN language model proposed outperforms the RNN proposed by \newcite{Mikolov:11} by about 10 points in accuracy. Furthermore, it improves upon the count-based dependency language model of \newcite{gubbins2013lm}, while achieving slightly worse than the recent state-of-the-art results by \newcite{mnih2013lblnce}. Finally, we make the code and preprocessed data available to facilitate comparisons with future work.

\section{Dependency Recurrent Neural Network}

Count-based language models operate by assigning probabilities to sentences by factorizing their likelihood into n-grams. Neural language models further \emph{embed} each word $w(t)$ into a low-dimensional vector representation (denoted by ${\bf s}(t)$)
\footnote{In our notation, we make a distinction between the word token $w(t)$ at position $t$ in the sentence and its one-hot vector representation ${\bf w}(t)$. We note $w_i$ the $i$-th word token on a breadth-first traversal of a dependency parse tree.}.

These word representations are learned as the language model is trained  \cite{bengio2003neural} and enable to define a word in relation to other words in a metric space.

\paragraph{Recurrent Neural Network}
\newcite{Mikolov:10} suggested the use of Recurrent Neural Networks (RNN) to model long-range dependencies between words as they are not restricted to a fixed context length, like the feedforward neural network \cite{bengio2003neural}.
The hidden  representation ${\bf s}(t)$ for the word in position $t$ of the sentence in the RNN follows a first order auto-regressive dynamic (Eq. \ref{eq:AR}), where ${\bf W}$ is the matrix connecting the hidden  representation of the previous word ${\bf s}(t-1)$ to the current one, ${\bf w}(t)$ is the one-hot index of the current word (in a vocabulary of size $N$ words) and ${\bf U}$ is the matrix containing the embeddings for all the words in the vocabulary: 
\begin{equation}
{\bf s}(t) = f \left ( {\bf W} {\bf s}(t-1) + {\bf U} {\bf w}(t) \right )
\label{eq:AR}
\end{equation}
The nonlinearity $f$ is typically the logistic sigmoid function $f(x) = \frac{1}{1 + \exp(-x)}$. 
At each time step, the RNN generates the word probability vector ${\bf y}(t)$ for the next word ${\bf w}(t+1)$, using the output word embedding matrix ${\bf V}$ and the softmax nonlinearity $g(x_i) = \frac{\exp(x_i)}{\sum _i \exp(x_i)}$:
\begin{equation}
{\bf y}(t) = g \left ( {\bf V} {\bf s}(t) \right )
\label{eq:out}
\end{equation}

\paragraph{RNN with Maximum Entropy Model}
\newcite{Mikolov:11} combined RNNs with a maximum entropy model, essentially adding a matrix that directly connects the input words' $n$-gram context ${\bf w}(t-n+1, \dots, t)$ to the output word probabilities. In practice, because of the large vocabulary size $N$, designing such a matrix is computationally prohibitive. Instead, a hash-based implementation is used, where the word context is fed through a hash function $h$ that computes the index $h({\bf w}(t-n+1, \dots, t))$ of the context words in a one-dimensional array ${\bf d}$ of size $D$ (typically, $D=10^9$). Array ${\bf d}$ is trained in the same way as the rest of the RNN model and contributes to the output word probabilities:
\begin{equation}
{\bf y}(t) = g \left ( {\bf V} {\bf s}(t) + {\bf d}_{h({\bf w}(t-n+1, \dots, t))} \right )
\label{eq:outDirect}
\end{equation}
As we show in our experiments, this additional matrix is crucial to a good performance on word completion tasks.

\paragraph{Training RNNs}
RNNs are trained using maximum likelihood through gradient-based optimization, such as Stochastic Gradient Descent (SGD) with an annealed learning rate $\lambda$. The Back-Propagation Through Time (BPTT) variant of SGD enables to sum-up gradients from consecutive time steps before updating the parameters of the RNN and to handle the long-range temporal dependencies in the hidden ${\bf s}$ and output ${\bf y}$ sequences. The loss function is the cross-entropy between the generated word distribution ${\bf y}(t)$ and the target \emph{one-hot} word distribution ${\bf w}(t+1)$, and involves the log-likelihood terms $\log y_{w(t+1)}(t)$. 

For speed-up, the estimation of the output word probabilities is done using hierarchical softmax outputs, i.e., class-based factorization \cite{Mikolov:12}. Each word $w^i$ is assigned to a class $c^i$ and the corresponding log-likelihood is effectively $\log y_{w^i}(t) = \log y_{c^i}(t) + \log y_{w^j}(t)$, where $j$ is the index of word $w^i$ among words belonging to class $c^i$. 
 In our experiments, we binned the words found in our training corpus into 250 classes according to frequency, roughly corresponding to the square root of the vocabulary size.

\paragraph{Dependency RNN}

RNNs are designed to process sequential data by iteratively presenting them with word ${\bf w}(t)$ and generating next word's probability distribution ${\bf y}(t)$ at each time step. They can be reset at the beginning of a sentence by setting all the values of hidden vector ${\bf s}(t)$ to zero.

\begin{figure}[t]
\centering
\begin{small}
  \includegraphics[width=1.0\columnwidth]{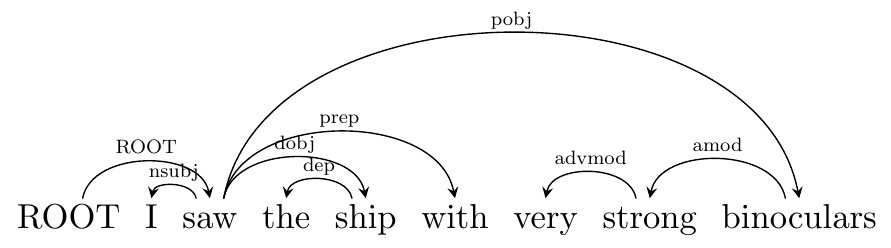}
\vspace{-0.2in}
\end{small}
\caption{Example dependency tree}
\label{fig:deps}
\end{figure}

Dependency parsing \cite{nivre2005dependency} generates, for each sentence (which we note $\{w(t)\}_{t=0}^T$), a parse tree with a single root, many leaves and an unique path (also called \emph{unroll}) from the root to each leaf, as illustrated on Figure~\ref{fig:deps}. 
We now note $\{w_i\}_i$ the set of word tokens appearing in the parse tree of a sentence. The order in the notation derives from the breadth-first traversal of that tree (i.e., the root word is noted $w_0$). Each of the unrolls can be seen as a different sequence of words $\{w_i\}$, starting from the single root $w_0$, that are visited when one takes a specific path on the parse tree.
We propose a simple transformation to the RNN algorithm so that it can process dependency parse trees. The RNN is reset and independently run on each such unroll.
As detailed in the next paragraph, when evaluating the log-probability of the sentence, a word token $w_i$ can appear in multiple unrolls but its log-likelihood is counted only once.
During training, and to avoid over-training the network on word tokens that appear in more than one unroll (words near the root appear in more unrolls than those nearer the leaves), each word token $w_i$ is given a weight discount $\alpha_i = \frac{1}{n_i}$, based on the number $n_i$ of unrolls the token appears in. Since the RNN is optimized using SGD and updated at every time-step, the contribution of word token $w_i$ can be discounted by multiplying the learning rate by the discount factor: $\alpha_i \lambda$.

\paragraph{Sentence Probability in Dependency RNN}

Given a word $w_i$, let us define the ancestor sequence $A(w_i)$ to be the subsequence of words, taken as a subset from $\{w_k\}^{i-1}_{k=0}$ and describing the path from the root node $w_0$ to the parent of $w_i$. For example, in Figure \ref{fig:deps}, the ancestors $A({\tt very})$ of word token {\tt very} are {\tt saw}, {\tt binoculars} and {\tt strong}.
Assuming that each word $w_i$ is conditionally independent of the words outside of its ancestor sequence, given its ancestor sequence $A(w_i)$, \newcite{gubbins2013lm} showed that the probability of a sentence (i.e., the probability of a lexicalized tree $S^T$ given an unlexicalized tree $T$) could be written as:
\begin{equation}
P [S^T | T] = \prod_{i=1}^{|S|}  P [w_i | A(w_i)]
\label{eq:ancestors}
\end{equation}

This means that the conditional likelihood of a word given its ancestors needs to be counted only once in the calculation of the sentence likelihood, even though each word can appear in multiple unrolls.
When modeling a sentence using an RNN, the state ${\bf s}_j$ that is used to generate the distribution of words ${\bf w}_i$ (where $j$ is the parent of $i$ in the tree), represents the vector embedding of the history of the ancestor words $A(w_i)$. Therefore, we count the term $P [{\bf w}_i | {\bf s}_j]$ only once when computing the likelihood of the sentence.

\section{Labelled Dependency RNN}

The model presented so far does not use dependency labels.
For this purpose we adapted the context-dependent RNN \cite{Mikolov:12} to handle them as additional $M$-dimensional label input features ${\bf f}(t)$. These features require a matrix ${\bf F}$ that connects label features to word vectors, thus yielding a new dynamical model (Eq. \ref{eq:ARcontext}) in the RNN, and a matrix ${\bf G}$ that connects label features to output word probabilities. The full model becomes as follows:
\begin{eqnarray}
\label{eq:ARcontext}
{\bf s}(t) & = & f \left ( {\bf W} {\bf s}(t-1) + {\bf U} {\bf w}(t) + {\bf F} {\bf f}(t) \right ) \\
\label{eq:outContext}
{\bf y}(t) & = & g \left ( {\bf V} {\bf s}(t) + {\bf G} {\bf f}(t) + {\bf d}_{h({\bf w}_{t-n+1}^t)} \right )
\end{eqnarray}
On our training dataset, the dependency parsing model found $M=44$ distinct labels (e.g., \emph{nsubj}, \emph{det} or \emph{prep}). At each time step $t$,  the context word ${\bf w}(t)$ is associated a single dependency label ${\bf f}(t)$ (a one-hot vector of dimension $M$).

Let $G(w)$ be the sequence of grammatical relations (dependency tree labels) between successive elements of $(A(w), w)$. The factorization of the sentence likelihood from Eq. \ref{eq:ancestors} becomes:
\begin{equation}
P [S^T | T] = \prod_{i=1}^{|S|}  P [w_i | A(w_i), G(w_i)] 
\end{equation}

\section{Implementation and Dataset}


We modified the Feature-Augmented RNN toolkit\footnote{ http://research.microsoft.com/en-us/projects/rnn/} and adapted it to handle tree-structured data. Specifically, and instead of being run sequentially on the entire training corpus, the RNN is run on all the word tokens in all unrolls of all the sentences in all the books of the corpus. The RNN is reset at the beginning of each unroll of a sentence. When calculating the log-probability of a sentence, the contribution of each word token is counted only once (and stored in a hash-table specific for that sentence). Once all the unrolls of a sentence are processed, the log-probability of the sentence is the sum of the per-token log-probabilities in that hash-table.
We also further enhanced the RNN library by replacing some large matrix multiplication routines by calls to the CBLAS library, thus yielding a two- to three-fold speed-up in the test and training time.\footnote{Our code and our preprocessed datasets are available from: \url{https://github.com/piotrmirowski/DependencyTreeRnn}}

The training corpus consists of 522 19th century novels from Project Gutenberg \cite{zweig2012challenge}. All processing (sentence-splitting, PoS tagging, syntactic parsing) was performed using the Stanford CoreNLP toolkit \cite{manning-EtAl:2014:P14-5}. The test set contains 1040 sentences to be completed. 
Each sentence consists of one ground truth and 4 impostor sentences where a specific word has been replaced with a syntactically correct but semantically incorrect \emph{impostor} word. Dependency trees are generated for each sentence candidate.
We split that set into two, using the first 520 sentences in the validation (development) set and the latter 520 sentences in the test set. During training, we start annealing the learning rate $\lambda$ with decay factor 0.66 as soon as the classification error on the validation set starts to increase.

\section{Results}



Table \ref{tab:seq} shows the accuracy (validation and test sets) obtained using a simple RNN with 50, 100, 200 and 300-dimensional hidden word representation and 250 frequency-based word classes (vocabulary size $N=72846$ words appearing at least 5 times in the training corpus). One notices that adding the direct word context to target word connections (using the additional matrix described in section 2), enables to jump from a poor performance of about 30\% accuracy to about 40\% test accuracy, essentially matching the 39\% accuracy reported for Good-Turing n-gram language models in \newcite{zweig2012computational}. Modelling 4-grams yields even better results, closer to the 45\% accuracy reported for RNNs in \cite{zweig2012computational}.\footnote{The paper did not provide details on the maximum entropy features or on class-based hierarchical softmax).}

As Table \ref{tab:dep} shows, dependency RNNs (depRNN) enable about 10 point word accuracy improvement over sequential RNNs.

\begin{table}[ht]
\begin{center}
\begin{tabular}{|l|cccc|}
\hline \bf Architecture & \bf 50h & \bf 100h & \bf 200h & \bf 300h \\
\hline
{\it RNN (dev)} & \it 29.6 & \it 30.0 & \it 30.0 & \it 30.6 \\
RNN (test) & 28.1 & 30.0 & 30.4 & 28.5 \\
\hline
{\it RNN+2g (dev)} & \it 29.6 & \it 28.7 & \it 29.4 & \it 29.8 \\
RNN+2g (test) & 29.6 & 28.7 & 28.1 & 30.2 \\
\hline
{\it RNN+3g (dev)} & \it 39.2 & \it 39.4 & \it 38.8 & \it 36.5 \\
RNN+3g (test) & 40.8 & 40.6 & 40.2 & 39.8 \\
\hline
{\it RNN+4g (dev)} & \it 40.2 & \it 40.6 & \it 40.0 & \it 40.2 \\
RNN+4g (test) & 42.3 & 41.2 & 40.4 & 39.2 \\
\hline
\end{tabular}
\end{center}
\caption{Accuracy of sequential RNN on the MSR Sentence Completion Challenge.}
\label{tab:seq}
\end{table}


\begin{table}[ht]
\begin{center}
\begin{tabular}{|l|ccc|}
\hline \bf Architecture & \bf 50h & \bf 100h & \bf 200h \\
\hline
{\it depRNN+3g (dev)} & 53.3 & \bf 54.2 & 54.2 \\
depRNN+3g (test) & 51.9 & \bf 52.7 & 51.9 \\
\hline
{\it ldepRNN+3g (dev)} & 48.8 & 51.5 & 49.0 \\
ldepRNN+3g (test) & 44.8 & 45.4 & 47.7 \\
\hline
{\it depRNN+4g (dev)} & 52.7 & 54.0 & 52.7 \\
depRNN+4g (test) & 48.9 & 51.3 & 50.8 \\
\hline
{\it ldepRNN+4g (dev)} & 49.4 & 50.0  & (48.5)  \\
ldepRNN+4g (test) & 47.7 & 51.4 & (47.7) \\
\hline
\end{tabular}
\end{center}
\caption{Accuracy of (un-)labeled dependency RNN (depRNN and ldepRNN respectively). }
\label{tab:dep}
\end{table}

The best accuracy achieved by the depRNN on the combined development and test sets used to report results in previous work was 53.5\%.
The best reported results in the MSR sentence completion challenge have been achieved by Log-BiLinear Models (LBLs) \cite{mnih2007lbl}, a variant of neural language models with  54.7\% to 55.5\% accuracy \cite{mnih2012fast,mnih2013lblnce}. We conjecture that their superior performance might stem from the fact that LBLs, just like n-grams, take into account the order of the words in the context and can thus model higher-order Markovian dynamics than the simple first-order autoregressive dynamics in RNNs.
The depRNN proposed ignores the left-to-right word order, thus it is likely that a combination of these approaches will result in even higher accuracies. \newcite{gubbins2013lm} developed a count-based dependency language model achieving 50\% accuracy. Finally, \newcite{mikolov2013efficient} report that they achieved 55.4\% accuracy with an ensemble of RNNs, without giving any other details.


\section{Discussion}

\paragraph{Related work}

\newcite{Mirowski:10} incorporated syntactic information into neural language models using PoS tags as additional input to LBLs but obtained only a small reduction of the word error rate in a speech recognition task. Similarly, \newcite{bian2014} enriched the Continuous Bag-of-Words (CBOW) model of \newcite{mikolov2013efficient} by incorporating morphology, PoS tags and entity categories into 600-dimensional word embeddings trained on the Gutenberg dataset, increasing sentence completion accuracy from 41\% to  44\%.
Other work on incorporating syntax into language modeling include \newcite{chelba1997structure} and \newcite{pauls2012large}, however none of these approaches considered neural language models, only count-based ones.
\newcite{levy2014embeddings} and \newcite{zhao2014} proposed to train neural word embeddings using skip-grams and CBOWs on dependency parse trees, but did not extend their approach to actual language models such as LBL and RNN and did not evaluate the word embeddings on word completion tasks.

Note that we assume that the dependency tree  is supplied prior to running the RNN which limits the scope of the Dependency RNN to the scoring of complete sentences, not to next word prediction (unless a dependency tree parse for the sentence to be generated is provided). Nevertheless, it is common in speech recognition and machine translation to use a conventional decoder to produce an N-best list of the most likely candidate sentences and then re-score them with the language model.   \cite{chelba1997structure,paulsfaster}

\newcite{tai2015} propose a similar approach to ours, learning Long Short-Term Memory (LSTM) \cite{hochreiter1997,graves2012} RNNs on dependency parse tree network topologies. Their architectures is not designed to predict next-word probability distributions, as in a language model, but to classify the input words (sentiment analysis task) or to measure the similarity in hidden representations (semantic relatedness task). Their relative improvement in performance (tree LSTMs vs standard LSTMs) on these two tasks is smaller than ours, probably because the LSTMs are better than RNNs at storing long-term dependencies and thus do not benefit form  the word ordering from dependency trees as much as RNNs.
In a similar vein to ours, \newcite{valerio2015} simply propose to enhance RNN-based machine translation by permuting the order of the words in the source sentence to match the order of the words in the target sentence, using a source-side dependency parsing.

\begin{figure}
\centering
  \includegraphics[width=1.0\columnwidth]{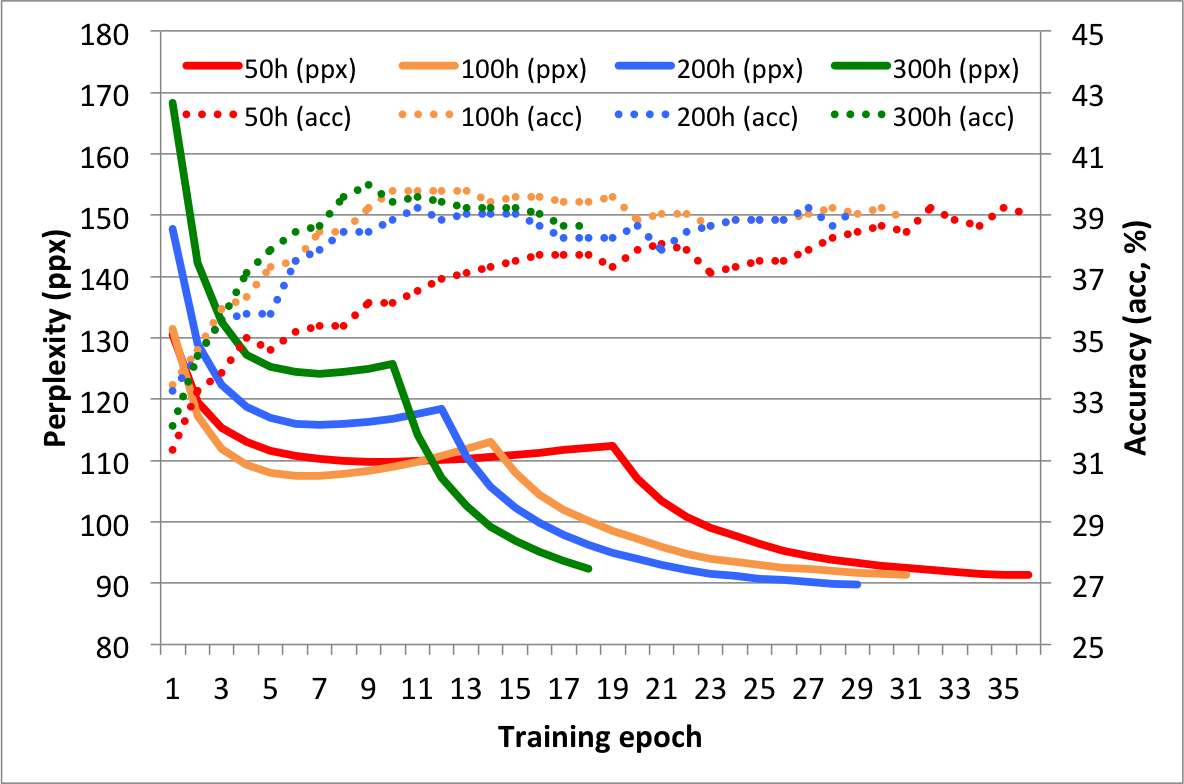}
  \caption{Perplexity vs. accuracy of RNNs}
\label{fig:test}
\end{figure}

\paragraph{Limitations of RNNs for word completion}
\newcite{zweig2012computational} reported that RNNs achieve lower perplexity than n-grams but do not always outperform them on word completion tasks.
As illustrated in Fig. \ref{fig:test}, the validation set perplexity (comprising all 5 choices for each sentence) of the RNN keeps decreasing monotonically 
 (once we start annealing the learning rate), whereas the validation accuracy rapidly reaches a plateau and oscillates. Our observation confirms that, once an RNN went through a few training epochs, change in perplexity is no longer a good predictor of change in word accuracy. We presume that the log-likelihood of word distribution is not a training objective crafted for $precision@1$, and that further perplexity reduction happens in the middle and tail of the word distribution.

\section{Conclusions}

In this paper we proposed a novel language model, dependency RNN, which incorporates syntactic dependencies into the RNN formulation. We evaluated its performance on the MSR sentence completion task and showed that it improves over RNN by 10 points in accuracy, while achieving results comparable with the state-of-the-art. Further work will include extending the dependency tree language modeling to Long Short-Term Memory RNNs to handle longer syntactic dependencies.

\section*{Acknowledgements}

We thank our anonymous reviewers for their valuable feedback. PM also thanks Geoffrey Zweig, Daniel Voinea, Francesco Nidito and Davide di Gennaro for sharing the original Feature-Augmented RNN toolkit on the Microsoft Research website and for insights about that code, as well as Bhaskar Mitra, Milad Shokouhi and Andriy Mnih for enlighting discussions about word embedding and sentence completion.

\bibliographystyle{acl}
\bibliography{MirowskiVlachosACL2015}

\end{document}